\documentclass[10pt,twocolumn,letterpaper]{article}

\usepackage{cvpr}              % To produce the CAMERA-READY version
\usepackage{subcaption}

\usepackage{graphicx} 
\usepackage{pifont}
\newcommand{\cmark}{\ding{51}}%
\newcommand{\xmark}{\ding{55}}%

\def\ie{{\textit{i.e.}}}
\def\eg{{\textit{e.g.}}}

\def\etc{{\textit{etc}}}

\usepackage{tablefootnote}
\newcommand{\tabincell}[2]{\begin{tabular}{@{}#1@{}}#2\end{tabular}}
\newcommand{\myPara}[1]{\vspace{0.1cm}\noindent\textbf{#1}}

\usepackage{wrapfig}
\usepackage{makecell}
\usepackage{amsthm,amsmath,amssymb}
\usepackage{mathrsfs}
\usepackage{arydshln} 
\usepackage{booktabs}
\usepackage{multirow}
\usepackage{caption}
\usepackage{amsfonts}
\usepackage{threeparttable}
\usepackage{tabularx}
\usepackage{longtable}
\usepackage{lipsum}
\usepackage{supertabular}
\usepackage{algorithm}
\usepackage{algorithmic}
\usepackage{subcaption}

\usepackage{wasysym} % 引入 wasysym 宏包
\definecolor{cvprblue}{rgb}{0.21,0.49,0.74}
\usepackage[pagebackref,breaklinks,colorlinks,allcolors=cvprblue]{hyperref}

\def\paperID{2} % *** Enter the Paper ID here
\def\confName{CVPR}
\def\confYear{2025}

\title{Underwater Camouflaged Object Tracking Meets Vision-Language SAM2}

\author{Chunhui Zhang$^{1,2,3}$, Li Liu$^{2,}$\thanks{Corresponding author: Li Liu (avrillliu@hkust-gz.edu.cn).}, Guanjie Huang$^{2}$, Zhipeng Zhang$^{1}$, Hao Wen$^{3}$, Xi Zhou$^{3}$,\\ Shiming Ge$^{4}$, Yanfeng Wang$^{1,5}$\\
$^{1}$Shanghai Jiao Tong University, Shanghai, 200240, China\\
$^{2}$The Hong Kong University of Science and Technology (Guangzhou), Guangzhou, 511458, China\\
$^{3}$CloudWalk Technology Co., Ltd, Shanghai, 201203, China\\
$^{4}$Institute of Information Engineering, Chinese Academy of Sciences, Beijing, 100085, China\\
$^{5}$Shanghai Artificial Intelligence Laboratory, Shanghai, 200032, China\\
}

\begin{document}
\maketitle
\begin{abstract}
Over the past decade, significant progress has been made in visual object tracking, largely due to the availability of large-scale datasets. However, these datasets have primarily focused on open-air scenarios and have largely overlooked underwater animal tracking—especially the complex challenges posed by camouflaged marine animals. To bridge this gap, we take a step forward by proposing the first large-scale multi-modal underwater camouflaged object tracking dataset, namely UW-COT220. Based on the proposed dataset, this work first comprehensively evaluates current advanced visual object tracking methods, including SAM- and SAM2-based trackers, in challenging underwater environments, \eg, coral reefs. Our findings highlight the improvements of SAM2 over SAM, demonstrating its enhanced ability to handle the complexities of underwater camouflaged objects. Furthermore, we propose a novel vision-language tracking framework called VL-SAM2, based on the video foundation model SAM2. Extensive experimental results demonstrate that the proposed VL-SAM2 achieves state-of-the-art performance across underwater and open-air object tracking datasets. The dataset and codes are available at~{\color{magenta}{https://github.com/983632847/Awesome-Multimodal-Object-Tracking}}.

%The dataset and codes are available at~\href{https://github.com/983632847/Awesome-Multimodal-Object-Tracking}{\color{magenta}{here}}.

\end{abstract}    
\section{Introduction}
\label{sec:intro}

Visual object tracking (VOT) involves continuously locating a target object within a video sequence and has applications in animal monitoring and conservation~\cite{katija2021visual,zhang2022webuav,zhang2023all}, autonomous vehicles~\cite{ye2022joint,ravi2024sam}, and surveillance~\cite{zhang2024awesome,fan2019lasot}, \etc. Its significance lies in enabling machines to perceive and interpret dynamic environments, supporting tasks like animal behavioral analysis~\cite{wang2023watb,chen2023alphatracker,zhang2024webuot} and decision-making~\cite{peng2024vasttrack,huang2019got}. Although substantial progress has been achieved in terrestrial and open-air scenarios~\cite{fan2021lasot,huang2019got,fan2019lasot}, tracking in underwater environments remains highly challenging, largely limiting the effectiveness of conventional algorithms~\cite{zhang2024webuot,alawode2023improving,alawode2022utb180}. Consequently, there is a pressing need for specialized datasets and methods to tackle the complexities of underwater tracking, especially when objects blend with their surroundings, known as camouflaged objects~\cite{bideau2016s,cheng2022implicit}. \emph{However, despite its importance, underwater camouflaged object tracking remains an unexplored field.}

Underwater object tracking presents numerous unique challenges due to light attenuation~\cite{zhang2024webuot,alawode2023improving}, scattering effects~\cite{alawode2022utb180,peng2024vasttrack}, and frequent camouflage scenarios~\cite{cheng2022implicit,bideau2016s}, which significantly degrade the performance of conventional tracking paradigms. Correlation filter-based methods~\cite{xu2019learning,zhang2019robust,ge2020cascaded,ge2019distilling,zhang2020accurate}, while computationally efficient, fail to handle the dynamic appearance variations caused by underwater optical distortions. Siamese networks~\cite{fan2019siamese,bertinetto2016fully,li2019siamrpn++,guo2022divert} struggle with feature representation learning in low-contrast and turbid environments. Transformer-based approaches~\cite{lin2024tracking,wu2023single,wang2021transformer,ye2022joint,chen2023seqtrack,wei2023autoregressive,zhang2025cost,zhang2024awesome}, though powerful in modeling long-range dependencies, often suffer from attention distraction due to cluttered underwater backgrounds. Even the emerging Mamba-based methods~\cite{huang2024mamba,lai2024mambavt,zhang2024mambatrack}, despite their efficient sequence modeling, lack the specialized capability to address underwater-specific degradations. In contrast, SAM- and SAM2-based trackers~\cite{zhang2023comprehensive,zhang2024segment,fu2023sam,yang2023track,yang2024samurai,videnovic2024distractor} demonstrate superior robustness by leveraging their pre-trained visual foundation model's strong generalization ability and prompt-guided segmentation mechanism. This enables effective handling of camouflaged marine animals through semantic-aware feature extraction and adaptive multi-scale processing, making SAM- and SAM2-based approaches particularly suitable for challenging underwater tracking scenarios.

\begin{figure*}[ht]
%\vspace{-0.5cm}
  \centering
  \includegraphics[width=1.0\textwidth]{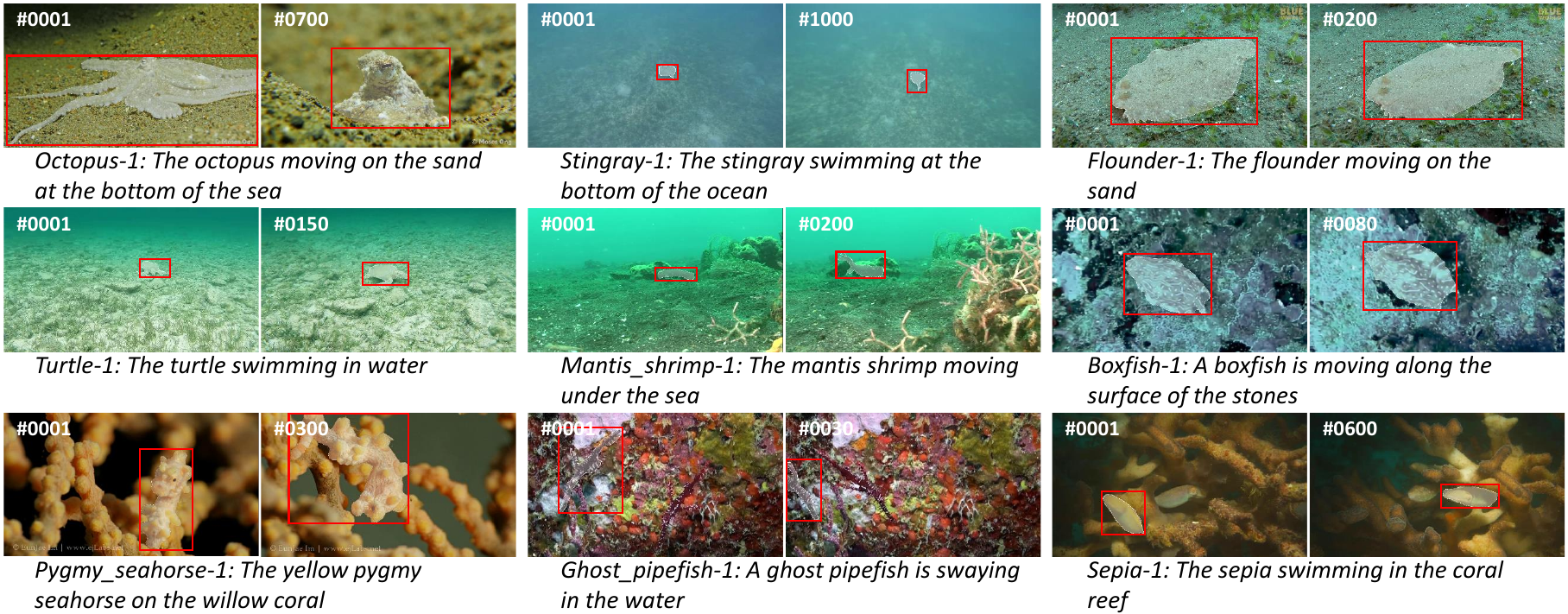}
  %\vspace{-0.6cm}
  \caption{A glance at some examples covering \emph{diverse underwater scenes} in our UW-COT220 dataset. We provide extensive annotations for underwater camouflaged target objects in video sequences, including \emph{bounding boxes}, \emph{masks}, and \emph{language descriptions}. }
  % Best viewed in color and zoomed in.
  \label{fig:examples}
%\vspace{-0.3cm}
\end{figure*}

\begin{table*}[ht]
{
%\scriptsize
\footnotesize
%\small
\renewcommand\arraystretch{1.0}
\caption{Comparison of the constructed UW-COT220 with existing camouﬂaged object tracking dataset (\ie, COTD~\cite{guo2024cotd}) and video camouflaged object detection/segmentation datasets (\ie, {CAD}~\cite{bideau2016s} and {MoCA-Mask}~\cite{cheng2022implicit}). UW-COT220 is the first multi-modal underwater camouflaged object tracking dataset.}
    %\vspace{-0.7cm}
	\label{tab:Comp_UWCOT}
	\begin{center}
		\setlength{\tabcolsep}{1.23mm}{
			\scalebox{1.0}{
			\begin{tabular}{lccccccccccccc}
			\Xhline{0.75pt} 
				Dataset & Year &  Videos & Classes  &   \tabincell{c}{ Min \\ frame} & \tabincell{c}{Mean\\ frame} & \tabincell{c}{Max\\frame} & \tabincell{c}{Total\\ frames} &  \tabincell{c}{Labeled \\frames}  &  \tabincell{c}{Visual \\annotation} & \tabincell{c}{Absent \\label}  & \tabincell{c}{Language \\descriptions}   & Underwater & \tabincell{c}{Open \\source}\\
				
			\hline
    
	        \textbf{CAD}~\cite{bideau2016s} &  2016  & 9 & 6  & 30 & 93 & 218 & 836   & 191 & Mask  & \xmark & \xmark  & \xmark  & \textcolor{green}{\scalebox{1.2}{\smiley}}\\
	        	
	        \textbf{MoCA-Mask}~\cite{cheng2022implicit} & 2022 & 87  & 45 &  23 & 264  & 1,296 & 23 K   & 23 K & Mask  & \xmark  & \xmark  & \xmark & \textcolor{green}{\scalebox{1.2}{\smiley}} \\
	        	
	        \textbf{COTD}~\cite{guo2024cotd} & 2024 & 200 & 20  & - & 400 & - & 80 K  & 80 K & BBox  & \xmark & \xmark  & \xmark  & \textcolor{red}{\scalebox{1.2}{\frownie}} \\
        	
		    \hline
	
		    \textbf{UW-COT220 (Ours)} & 2025 &  220 & 96  & 10  & 722 & 7,448 &  159 K & 159 K &  BBox+Mask  & \cmark & \cmark & \cmark &  \textcolor{green}{\scalebox{1.2}{\smiley}}\\
		    \Xhline{0.75pt} 
	\end{tabular}
	   }}
\end{center}}
%\vspace{-0.6cm}
\end{table*}

Recently, several video camouflaged object detection and segmentation datasets have been proposed~\cite{bideau2016s,cheng2022implicit,guo2024cotd}, but they primarily focus on open-air scenarios. Following the latest large-scale underwater object tracking dataset, \ie, WebUOT-1M~\cite{zhang2024webuot}, we take a step forward and construct the first multi-modal underwater camouflaged object tracking dataset, UW-COT220 (see Fig.~\ref{fig:examples} and Tab.~\ref{tab:Comp_UWCOT}), as a benchmark for evaluating and developing advanced underwater camouflaged object tracking methods. The dataset consists of 220 underwater video sequences, spanning 96 categories, with approximately 159 K frames. We annotate each video with a language description to promote the exploration of multi-modal underwater camouflaged object tracking. As a single-object visual tracking task, we provide bounding box annotations for the camouflaged underwater objects in each frame. To support precise tracking, we annotate each frame with absent labels: ``0'' for target presence and ``1'' for full occlusion or out-of-view. Additionally, using manually annotated bounding boxes as prompts, we generate mask annotations for the camouflaged underwater objects. To ensure the quality of the annotations, we have carefully checked the bounding boxes, masks, and language descriptions for multiple rounds. To advance research in the related field, we propose a simple yet effective VL tracking framework, called VL-SAM2. By releasing our dataset and codes, we aim to inspire the exploration of diverse underwater vision tasks and foster the development of underwater camouflaged object tracking and beyond.

Our main contributions can be summarized as follows:
\begin{itemize}
  \item We introduce UW-COT220, the first large-scale multi-modal underwater camouflaged object tracking dataset with bounding boxes, masks, and language descriptions.
  \item We propose VL-SAM2, a new VL tracker based on SAM2, leveraging language prompts and motion-aware prediction for robust underwater tracking.
  \item Our VL-SAM2 outperforms state-of-the-art (SOTA) tracking methods, proving its effectiveness in handling underwater camouflage and complex scenes.
\end{itemize}

%\vspace{-0.2cm}
\section{UW-COT220 Dataset}
\label{sec:section2}
%\vspace{-0.2cm}

\begin{figure*}[t]
%\vspace{-0.5cm}
  \centering
  \includegraphics[width=1.0\textwidth]{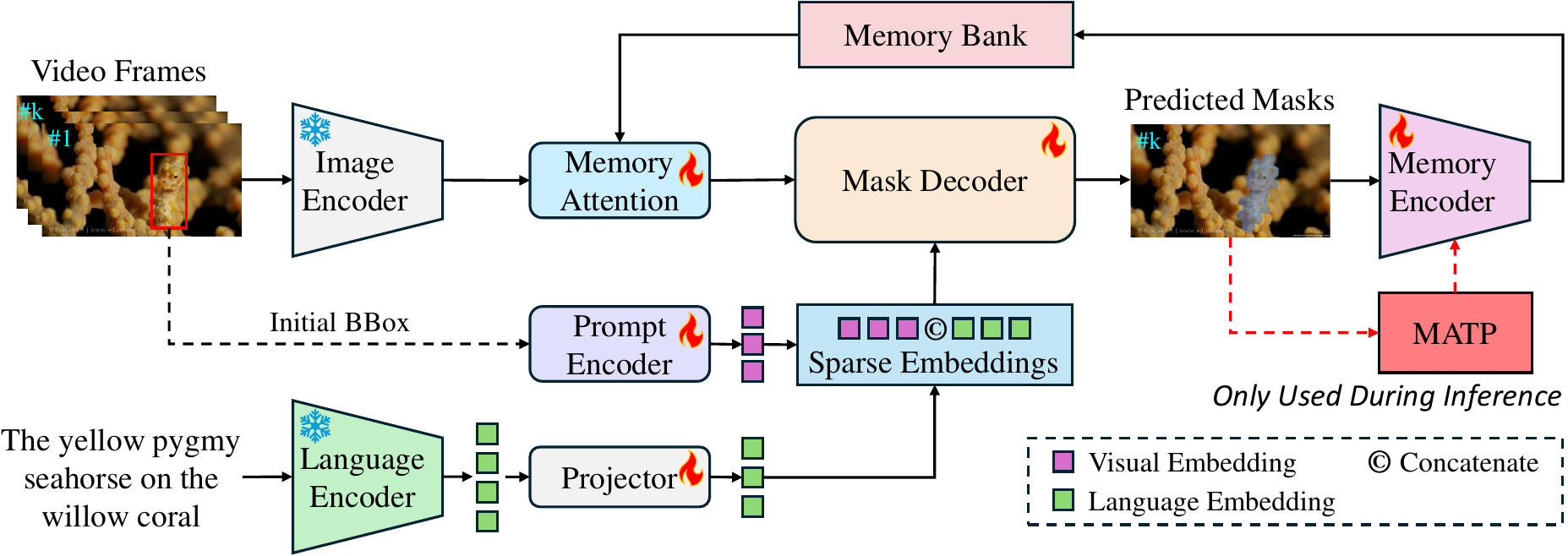}
  %\vspace{-0.3cm}
  \caption{Overview of the proposed vision-language tracking framework VL-SAM2. Our VL-SAM2 retains the core architecture of SAM2 and freezes the weights of the image encoder. The language prompt is encoded by the language encoder and then injected into the mask decoder. Additionally, a \emph{training-free} MATP module is integrated to enhance tracking robustness during inference.}
  \label{fig:vlsam2}
%\vspace{-0.4cm}
\end{figure*}

Our goal is to construct a large-scale underwater camouflaged object tracking dataset that involves a wide variety of categories and various real underwater scenes for evaluating and developing general underwater camouflaged object tracking methods. To achieve this, we collect underwater videos from video-sharing platforms (\eg, YouTube\footnote{https://www.youtube.com/}) and existing tracking datasets (\eg, WebUOT-1M~\cite{zhang2024webuot} and VastTrack~\cite{peng2024vasttrack}), and we filter out 220 videos that contain underwater camouflaged objects. Note that WebUOT-1M~\cite{zhang2024webuot} is currently the largest general-purpose underwater object tracking dataset, comprising 81 underwater camouflage videos. We further construct a large-scale dataset dedicated to underwater camouflaged object tracking, featuring more diverse underwater scenes and comprehensive annotations. Following existing datasets~\cite{zhang2022webuav,peng2024vasttrack,zhang2024webuot}, we provide a bounding box annotation for the camouflaged object in each frame, represented as $[x, y, w, h]$, where $x$ and $y$ denote the coordinates of the top-left corner of the object, and $w$ and $h$ indicate the width and height of the object. In each video, we annotate an English sentence describing the color, behavior, attributes, and surroundings of the target. Furthermore, to enhance the accuracy of object representation, we use the SAM-powered interactive semi-automatic annotation tool~\cite{ISAT_with_segment_anything} to generate mask annotations for the camouflaged objects. Some examples and detailed statistics of the UW-COT220 dataset are provided in Fig.~\ref{fig:examples} and Tab.~\ref{tab:Comp_UWCOT}. To the best of our knowledge, UW-COT220 is \emph{\textbf{the first large-scale multi-modal benchmark for underwater camouflaged object tracking}}, featuring a diverse set of 96 categories and various challenging underwater scenes.

%\vspace{-0.3cm}
\section{Proposed Method}
\label{sec:section3}
%\vspace{-0.2cm}

\begin{algorithm}[t]
%\scriptsize
\footnotesize
%\small	\renewcommand{\algorithmicrequire}{\textbf{Input:}}
	\renewcommand{\algorithmicensure}{\textbf{Output:}}
	\caption{ \small Inference with MATP}
	\label{alg:alg1}
        \begin{algorithmic}[1]
            \REQUIRE Kalman filter $kf$, first frame, initial box, response map, candidate set $C_{t}$, maximum response set $C'_{t}$, scores list $scores$, estimation box $b^{E}$, match result box $b^{M}$, IoU threshold $conf = 0.6$, response map threshold $threshold = 0.8$, IoU threshold $iou\_threshold = 0.5$, match state $match\_state = False$.
            
            \ENSURE Target boxes ${B}$
            \STATE $kf$.init(first frame)
            \STATE $B$=[initial box]
            
            \FOR{$i=2,3,...,T$ frames}
                    \STATE $C_{t} \leftarrow$ extract\_candidates(response map, $threshold$)
                    \STATE $C'_{t} \leftarrow$ NMS($C_{t}$)
                    \STATE $b^{E} \leftarrow kf$.predict()
                    \STATE $scores \leftarrow$ compute\_scores($b^{E}$, $C'_{t}$)
            
                    \IF{iou\_of(max response box, $b^{E}$) $< conf$}
                        \STATE $match\_state \leftarrow True$
                    \ELSE
                        \STATE $match\_state \leftarrow False$
                    \ENDIF
            
                    \IF{$match\_state$}
                        \STATE $b^{M} \leftarrow$ argmax($scores$)
                    \ELSE
                        \STATE $b^{M} \leftarrow$ max response box
                    \ENDIF
                    
                \STATE $B$.append($b^{M}$)
                \STATE $kf$.update($b^{M}$)

            \ENDFOR
            \RETURN $B$
            
        \end{algorithmic}
       
\end{algorithm}

In this section, we introduce vision-language (VL)-SAM2, a new vision-language tracking framework for multi-modal underwater camouflaged object tracking. The proposed VL-SAM2 (see Fig.~\ref{fig:vlsam2}) is based on the latest video foundation model, SAM2~\cite{yang2024samurai}, and extends its capabilities to underwater scenarios. VL-SAM2 consists of three main components, including a visual branch (\ie, SAM2), a language branch (LB), and a motion-aware target prediction (MATP) module~\cite{zhang2024webuot}. 

\begin{figure*}[t]
%\vspace{-0.5cm}
    \centering
    \begin{subfigure}{0.245\textwidth}
        \includegraphics[width=\textwidth]{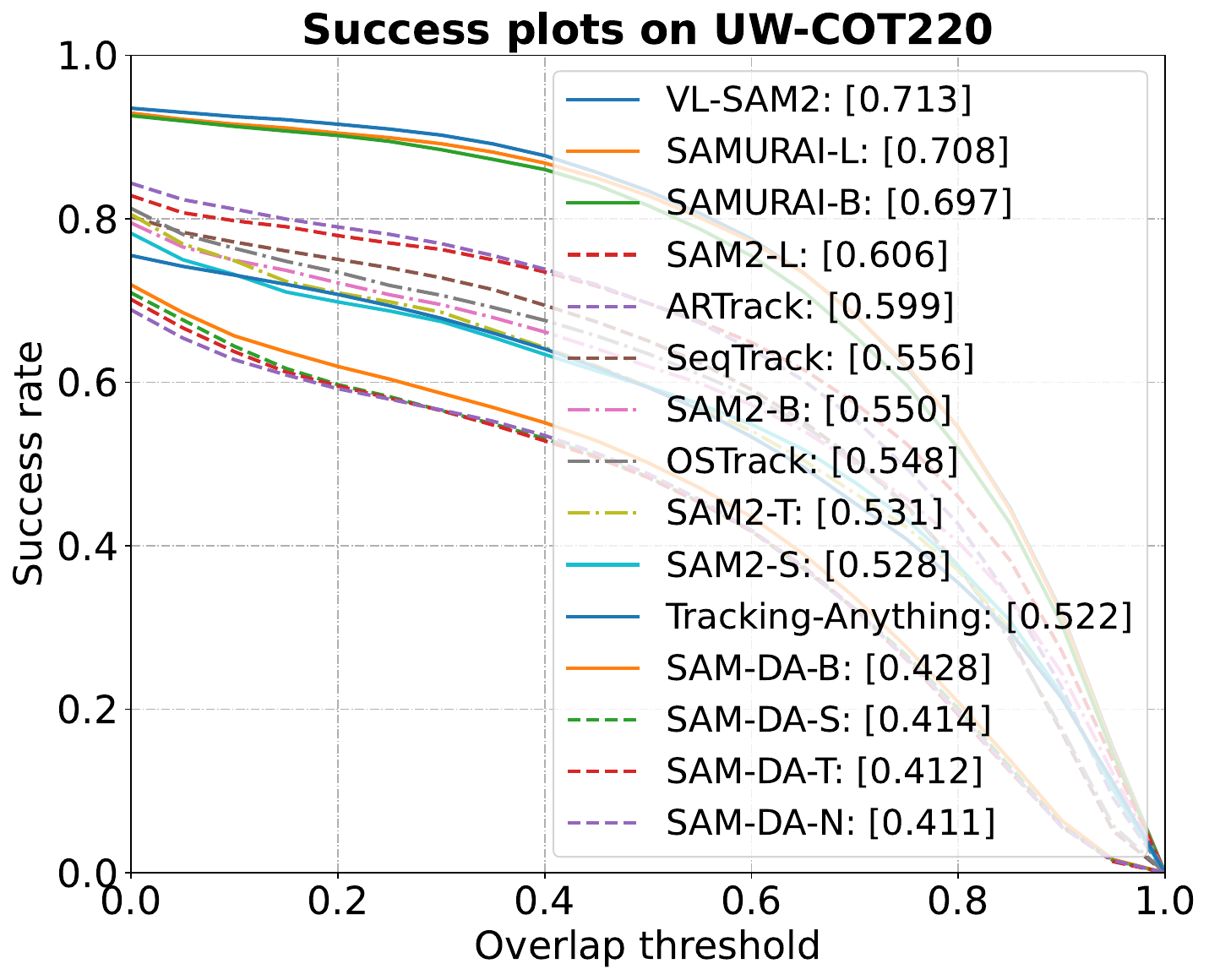}
    \end{subfigure}
    \begin{subfigure}{0.245\textwidth}
        \includegraphics[width=\textwidth]{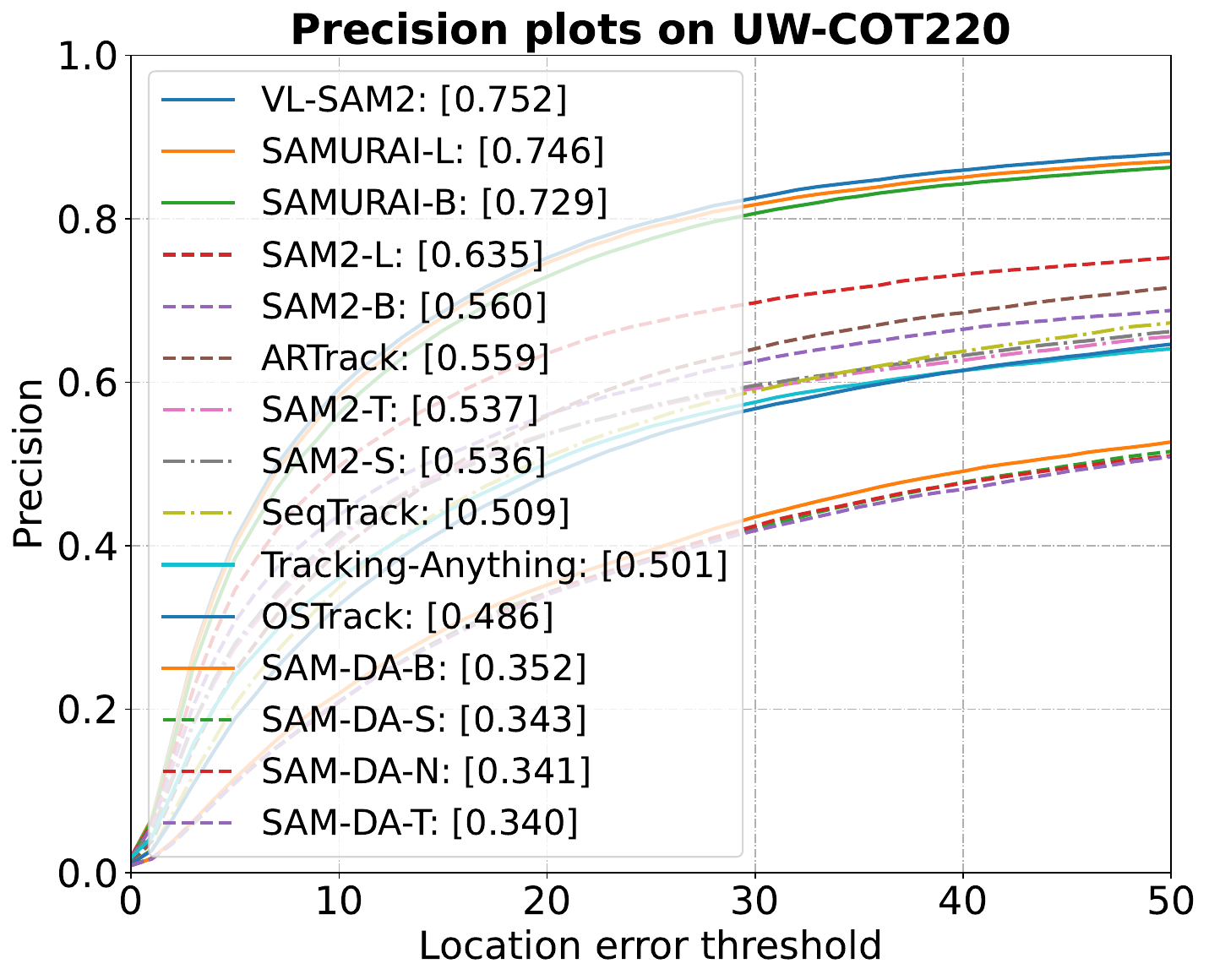}
    \end{subfigure}
    \begin{subfigure}{0.245\textwidth}
        \includegraphics[width=\textwidth]{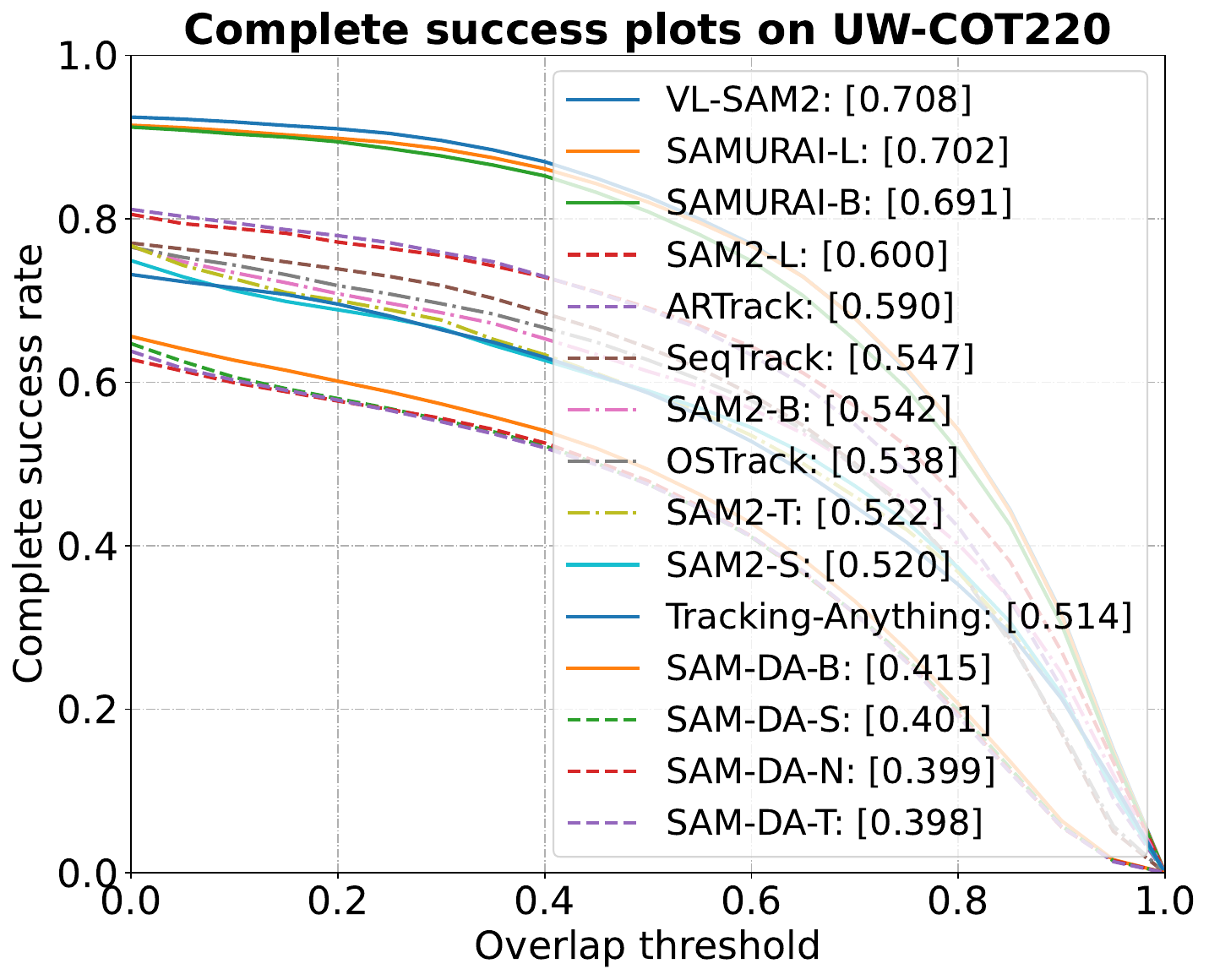}
    \end{subfigure}
    \begin{subfigure}{0.245\textwidth}
        \includegraphics[width=\textwidth]{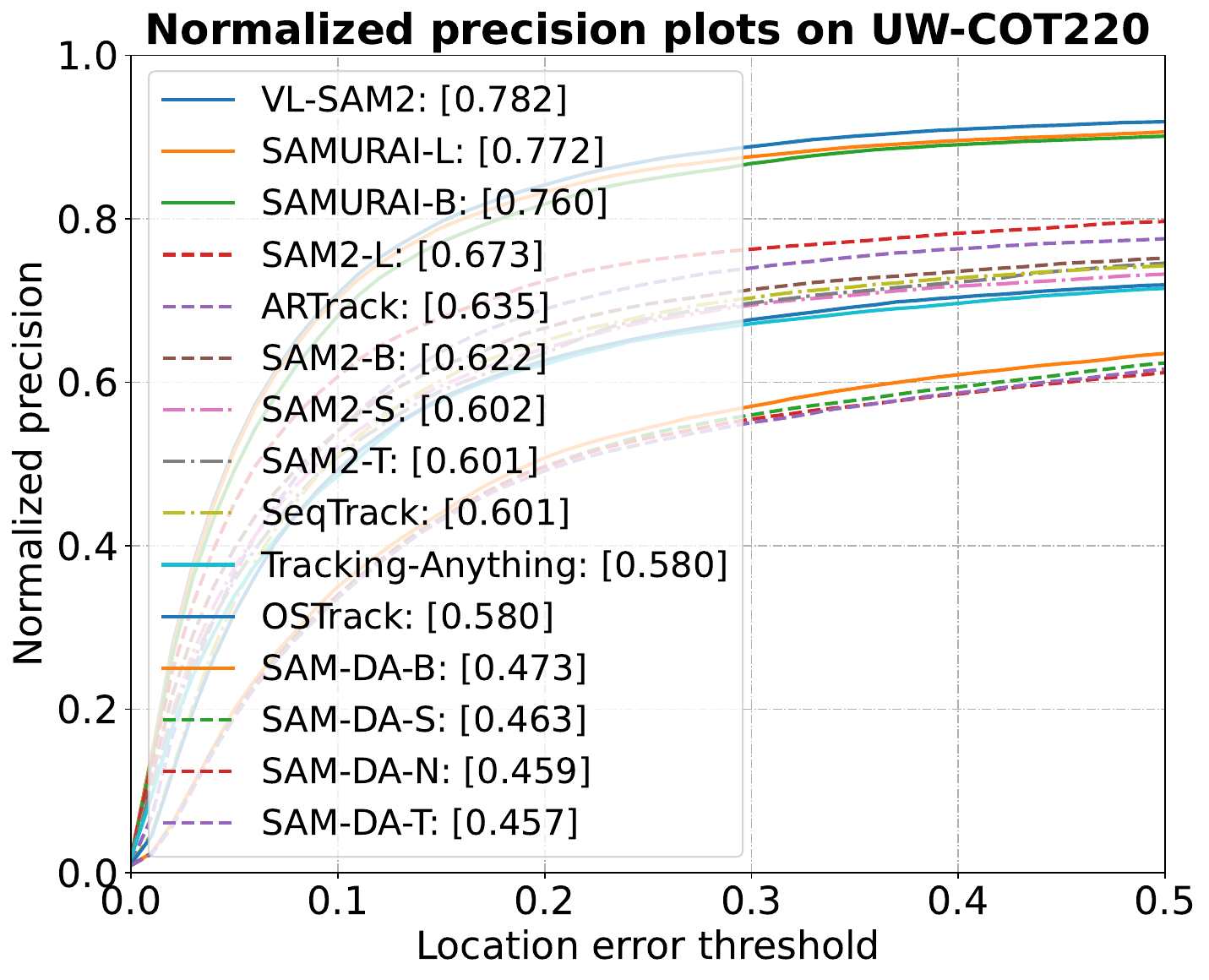}
    \end{subfigure}
    %\vspace{-0.3cm}
    \caption{Comparison of SOTA trackers on the UW-COT220 dataset using AUC, Pre, cAUC, and nPre scores. Best viewed by zooming in.}
    \label{fig:four_images}
    %\vspace{-0.4cm}
\end{figure*}

The input to VL-SAM2 consists of video frames and a language description. First, we use an image encoder and a language encoder to extract features from both modalities. In the first frame, the initial bounding box is encoded by the prompt encoder to obtain the visual embedding. The language embedding is processed by a linear projection to match the dimension of the visual embedding. Then, we concatenate the visual and language embeddings as sparse embeddings and feed them into the mask decoder. In subsequent frames, we adopt a video propagation manner similar to SAM2 to obtain the prediction result for each frame. The MATP, based on Kalman filtering~\cite{kalman1960new}, is used to alleviate the model drift caused by similar distractors around the true target. Its core idea is to predict the current state (\ie, target box) from the previous state and then refine it by integrating current observations to determine the optimal state. Algorithm~\ref{alg:alg1} presents the pseudo-code for MATP, which serves as a plug-and-play module to refine the predicted bounding boxes for more precise target boxes in subsequent frames. Inspired by~\cite{yang2024samurai}, we further apply MATP to update the memory of SAM2.

We use Hiera-L~\cite{ryali2023hiera} as the image encoder and CLIP ViT-B/32~\cite{radford2021learning} as the language encoder. During training, we freeze the parameters of both the image encoder and the language encoder, and fine-tune the other modules. This allows us to train the proposed VL-SAM2 on an Ubuntu server with 8 A6000 GPUs. Our model was trained on the Ref-YouTube-VOS~\cite{seo2020urvos} dataset and evaluated in a \emph{zero-shot} manner on the UW-COT220 and existing datasets~\cite{fan2019lasot,zhang2024webuot}.

%%%%%%%%%%%%%%%%%%%%%%%%%%%%%%%%%%%%%%%%%%%%%%%%%%%%%%%%%%%%
%\vspace{-0.3cm}
\section{Experiments}
\label{sec:section4}

\myPara{Main Results.} We evaluate SAM/SAM2-based tracking methods (\ie, SAM-DA~\cite{fu2023sam}, Tracking Anything~\cite{yang2023track}, SAM2, SAMURAI-L~\cite{yang2024samurai}, and VL-SAM2), and current SOTA VOT methods (\ie, OSTrack~\cite{ye2022joint}, SeqTrack~\cite{chen2023seqtrack}, and  ARTrack~\cite{wei2023autoregressive}) on the proposed UW-COT220. The results are shown in Fig.~\ref{fig:four_images}. Our observations are as follows: 1) Our VL-SAM2 achieves the best performance, surpassing the SOTA trackers, including SAMURAI-L, SeqTrack and ARTrack. 2) The top three trackers (VL-SAM2, SAMURAI-L, and SAM2) are all SAM2-based, highlighting the promising trend of using video foundation models for tracking. 3) SAM2-based methods outperform SAM-based trackers (SAM-DA and Tracking-Anything) on UW-COT220, which can be attributed to a series of improvements SAM2 introduces over SAM for video and image tasks, such as improving temporal consistency, robustness to occlusions, feature embeddings, computational efficiency, motion estimation accuracy, generalization to new domains, and integration of contextual information~\cite{ravi2024sam,zhang2024segment}.

\myPara{Impact of Prompt.} We take SAM2~\cite{ravi2024sam} as our baseline tracker to explore the impact of different ways of point prompts (\ie, center point and random point within the initial target box). From Tab.~\ref{tab:point_prompt}, we find that using the center point as a prompt yields significantly better results than using random points. This implies that for the interactive segmentation model SAM2, the quality of the input prompt plays a crucial role in determining its performance. This also motivates us to explore the use of more flexible prompts, \eg, language descriptions, in the context of underwater camouflaged object tracking.

\begin{table}[t]
%\vspace{-0.2cm}
\footnotesize
  \centering
  \caption{Impact of using different points (within the initial target box) as the point prompt for SAM2. We report AUC, nPre, Pre, cAUC, and mACC scores on the UW-COT220 dataset.}
  %\vspace{-0.3cm}
  \label{tab:point_prompt}
  \setlength{\tabcolsep}{1.8mm}{
  \scalebox{1.0}{
  \begin{tabular}{lcccccccc}
    \Xhline{0.75pt} % 表格横线加粗
    Method   &  Point type & AUC & Pre   & nPre  & cAUC & mACC \\
    \hline
    %\multicolumn{1}{l}{\multirow{2}[1]{*}{SAM2-T}} & Center point   & 53.1   &   53.7 &  60.1  &  52.2    &  53.4    \\
    % & Random point   &  37.7   &   37.2 &  41.6 &  36.4   &  37.3    \\
    %\hline
    
    %\multicolumn{1}{l}{\multirow{2}[1]{*}{SAM2-S}} &  Center point   & 52.8   &   53.6 &  60.2   &  52.0  &  53.1    \\
    % &  Random point   & 36.3  &   35.3 &  41.0  &  35.2  &    36.0  \\
    %\hline
    
    \multicolumn{1}{l}{\multirow{2}[1]{*}{SAM2-B}}  &	Center point  & 55.0   &   56.0 &  62.2  &  54.2   &  55.4    \\
      &	Random point  &  40.5   &   40.4 &  45.8  &  39.5  &  40.2   \\

    \hline
    \multicolumn{1}{l}{\multirow{2}[1]{*}{SAM2-L}}   &	Center point  & 60.6   &  63.5 &  67.3  &  60.0  &  61.3   \\
      &	Random point  & 44.6   &   45.5 &  49.3  &  43.5   &  44.6    \\
    
    \Xhline{0.75pt} % 表格横线加粗
  \end{tabular}
  }}
\end{table}

\begin{table}[t]
%\vspace{-0.2cm}
\footnotesize
  \centering
  \caption{Comparison of different models for SAM2 on the UW-COT220 dataset.}
  \label{tab:model_size}
  %\vspace{-0.3cm}
  \setlength{\tabcolsep}{1.1mm}{
  \scalebox{1.0}{
  \begin{tabular}{lcccccccc}
    \Xhline{0.75pt} % 表格横线加粗
    Method   & Size (M) &  Speed (FPS)  & AUC & Pre   & nPre  & cAUC & mACC \\
    \hline
    SAM2-T & 38.9 & 	47.2 & 53.1   &   53.7 &  60.1  &  52.2    &  53.4    \\

    SAM2-S  &	46.0	& 43.3  & 52.8   &   53.6 &  60.2   &  52.0  &  53.1    \\
    
    SAM2-B  &	80.8	& 34.8 &  55.0   &   56.0 &  62.2  &  54.2   &  55.4    \\
    
    SAM2-L  &	224.4	& 24.2 &  60.6   &  63.5 &  67.3  &  60.0  &  61.3   \\

    \Xhline{0.75pt} % 表格横线加粗
  \end{tabular}
  }}
  %\vspace{-0.33cm}
\end{table} 

\begin{table}[t]
\footnotesize
  \centering
  \caption{Impact of different language encoders on the UW-COT220 dataset.}
  %\vspace{-0.3cm}
  \label{tab:langauge_encoder}
  \setlength{\tabcolsep}{1.3mm}{
  \scalebox{1.0}{
  \begin{tabular}{lcccccccc}
    \Xhline{0.75pt} % 表格横线加粗
    Method   & Language Encoder & AUC & Pre   & nPre  & cAUC & mACC\\
    \hline
    VL-SAM2 & T5~\cite{raffel2020exploring} & 54.6   &  54.8  &  59.3  &  53.9    &  55.1   \\

    VL-SAM2  &	CLIP~\cite{radford2021learning}  & 71.3    &   75.2   &  78.2  & 70.8  &  72.4    \\
    
    \Xhline{0.75pt} % 表格横线加粗
  \end{tabular}
  }}
\end{table}

\myPara{Model Size \& Speed Analysis.} Tab.~\ref{tab:model_size} demonstrates that larger model sizes generally lead to better performance, but the speed of the model decreases significantly. We also discovered an interesting phenomenon: when the number of model parameters is relatively small, a smaller model (\eg, SAM2-T) can even outperform a larger model (\eg, SAM2-S). We suspect this may be due to overfitting, or that relatively small models (\eg, SAM2-S) are more sensitive to the quality of the training data~\cite{ravi2024sam}.

\myPara{Impact of Language Encoder.} In Tab.~\ref{tab:langauge_encoder}, we explore the impact of using different language encoders in the proposed VL-SAM2. T5~\cite{raffel2020exploring} is a large language model that models natural language processing tasks as text-to-text problems, while CLIP~\cite{radford2021learning} is a contrastive language-image pre-training model. The result demonstrates that using CLIP as the language encoder outperforms T5. We hypothesize that this is due to CLIP's text encoder being aligned with the visual encoder in the feature space over a vast amount of data, making it more suitable for the VOT task.

\myPara{Generalization Analysis.} 
To demonstrate the generalization ability of the proposed VL-SAM2, we report the results on the large-scale open-air tracking dataset, \ie, LaSOT~\cite{fan2019lasot}, and the generic underwater object tracking benchmark, \ie, WebUOT-1M~\cite{zhang2024webuot} (see Tab.~\ref{tab:more_results}). Results show that VL-SAM2 outperforms the current SOTA trackers. We believe that VL-SAM2 demonstrates strong generalization capability, ranging from challenging \emph{underwater scenarios} to complex \emph{open-air environments}.

\begin{figure}[t]
\centering
\includegraphics[width=0.48\textwidth]{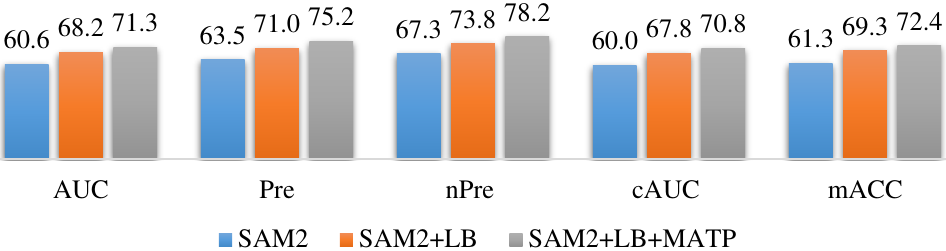}
%\vspace{-0.6cm}
  \caption{Ablation study on the UW-COT220 dataset.}
  \label{fig:ablation_study}
%\vspace{-0.6cm}
\end{figure}

\myPara{Ablation Study.} We conduct experiments to verify two core modules in the VL-SAM2 framework, \ie, the language branch (LB) and MATP. As shown in Fig.~\ref{fig:ablation_study}, injecting language prompts via LB significantly improves performance compared to the baseline tracker. The performance is further enhanced by using the MATP module.

\begin{table}[t]
%\scriptsize
\footnotesize
%\vspace{-0.2cm}
  \caption{Evaluation on LaSOT and WebUOT-1M datasets.}
  %\vspace{-0.3cm}
  \label{tab:more_results}
  \centering
  \setlength{\tabcolsep}{1.9mm}{
  \scalebox{1.0}{
  \begin{tabular}{lccccccc}
    \Xhline{0.75pt} % 表格横线加粗
      \multicolumn{1}{l}{\multirow{2}[1]{*}{Method}} &  \multicolumn{3}{c}{ LaSOT }  & \multicolumn{3}{c}{  WebUOT-1M} \\
   
   \cmidrule(r){2-4} \cmidrule(r){5-7}  &  AUC  &  Pre & nPre & AUC  &  Pre  & nPre\\
    \hline
          
    %JointNLT~\cite{zhou2023joint}  &   60.4 & 63.6 & 69.4  &  xxx   &   xxx  &  xxx   \\

    CiteTracker-256~\cite{li2023citetracker}   &   65.9 & 70.2 & 75.7  &  54.6   &   49.3  &  58.4   \\
    
    VLT$_{\rm TT}$~\cite{guo2022divert}    &   67.3 &   72.1   &   77.6  &  48.3   &   41.7  &  52.1   \\

    OSTrack~\cite{ye2022joint}  &   69.1 & 75.2 & 78.7  &  56.5   &   52.9  &  61.1   \\
    
    MixFormerV2-B~\cite{cui2023mixformerv2}   & 69.5  & 75.0 &  79.1 &  51.0   &   45.4  &  54.0   \\
    
    SeqTrack-B256~\cite{chen2023seqtrack}  &   69.9 & 76.3 & 79.7  &  54.0   &   50.8  &  58.5   \\

    %All-in-One~\cite{zhang2023all}   &   71.7 &   78.5   &   82.4  &  57.1   &   53.1  &  61.5   \\
    \hline
    
    SAM2-L~\cite{ravi2024sam}   &   59.3 &   61.9   &   64.5  &  60.6   &   65.6  &  69.2   \\
    
    VL-SAM2 (Ours)   &  \textbf{73.4} &   \textbf{79.4}   &   \textbf{81.8}  &  \textbf{71.9}   &  \textbf{77.2}  &  \textbf{80.0}   \\

    \Xhline{0.75pt} % 表格横线加粗

  \end{tabular}
  }}
  %\vspace{-0.3cm}
\end{table}

\section{Conclusion}
\label{sec:section5}
%\vspace{-0.2cm}
In this work, we introduce UW-COT220, the first large-scale multi-modal underwater camouflaged object tracking dataset. To promote related research, we propose a new vision-language tracking framework VL-SAM2. Experimental results show that our proposed VL-SAM2 exhibits excellent zero-shot generalization capability in both underwater and open-air scenarios. Future work includes expanding the scale and modalities of our dataset and exploring more underwater vision tasks.

\myPara{Acknowledgements.} This work was supported by the National Natural Science Foundation of China (No. 62471420), GuangDong Basic and Applied Basic Research Foundation (2025A1515012296), CCF-Tencent Rhino-Bird Open Research Fund, and the Major Project of Technology Innovation and Application Development of Chongqing (CSTB2023TIAD-STX0015).

{   
    %\vspace{-0.3cm}
    \small
    \bibliographystyle{ieeenat_fullname}
    \bibliography{main}
}


\end{document}